# SSAS: Cross-subject EEG-based Emotion Recognition through Source Selection with Adversarial Strategy

Yici Liu[a], Qi Wei Oung[b] and Hoi Leong Lee[b,*]

[a]School of Computer Science and Engineering, Southeast University, 210096, Nanjing, China.
[b]Faculty of Electronic Engineering and Technology, Universiti Malaysia Perlis, 02600 Arau, Perlis, Malaysia.

## ABSTRACT

Electroencephalographic (EEG) signals have long been applied in the field of affective brain-computer interfaces (aBCIs). Cross-subject EEG-based emotion recognition has demonstrated significant potential in practical applications due to its suitability across diverse people. However, most studies on cross-subject EEG-based emotion recognition neglect the presence of inter-individual variability and negative transfer phenomena during model training. To address this issue, a cross-subject EEG-based emotion recognition through source selection with adversarial strategy is introduced in this paper. The proposed method comprises two modules: the source selection network (SS) and the adversarial strategies network (AS). The SS uses domain labels to reverse-engineer the training process of domain adaptation. Its key idea is to disrupt class separability and magnify inter-domain differences, thereby raising the classification difficulty and forcing the model to learn domain-invariant yet emotion-relevant representations. The AS gets the source domain selection results and the pretrained domain discriminators from SS. The pretrained domain discriminators computes a novel loss aimed at enhancing the performance of domain classification during adversarial training, ensuring the balance of adversarial strategies. This paper provides theoretical insights into the proposed method and achieves outstanding performance on two EEG-based emotion datasets, SEED and SEED-IV. The code can be found at https://github.com/liuyici/SSAS.

## 1. Introduction

Due to its ability to record neuronal electrical activity at different times and locations, electroencephalographic (EEG) signals offer advantages such as repeatability, non-invasiveness, and high temporal resolution. These attributes have led to broader attention for EEG-based emotion recognition [1, 2, 3]. Various learning approaches have been applied in this field, including supervised learning, which relies on labeled data but faces challenges in obtaining large, annotated EEG datasets, and unsupervised learning, which circumvents labeling but may struggle with accuracy in emotion classification [4]. Few-shot learning methods address data scarcity by learning from minimal labeled samples, however, target domain labels are difficult to obtain in some scenarios. Cross-subject transfer learning approaches, which leverage data from multiple subjects, have been widely adopted to reduce the annotation cost in practical applications [5]. Cross-subject EEG-based emotion recognition does not require labeled target domain and is characterized by strong generalization capability. Nevertheless, two key challenges remain. Firstly, a key problem in cross-subject EEG-based emotion recognition is the inter-individual variability, which complicates the construction of a model with robust generalization capability. Secondly, negative transfer phenomena induced by training data: due to the variability in the quality of source data, which may be caused by patients' unfamiliarity with the EEG acquisition process, certain data instances may weaken the model's generalization capability.

In recent years, domain adaptation (DA) have emerged as powerful techniques in EEG-based emotion recognition. These methods can be broadly categorized into distribution alignment-based methods, graph neural network-based methods, adversarial learning-based methods, source selection-based methods, and subspace mapping-based methods. In distribution alignment-based methods, Pan et al. [6] were the first to employ Maximum Mean Discrepancy (MMD) to reduce the distribution differences between the source and target domains. Zhu et al. [7] utilized the Wasserstein distance as a metric for DA, while Zhu et al. [8] minimized distribution differences by considering domain-specific decision boundaries between classes. Luo et al. [9] introduced Wasserstein generative adversarial network with gradient penalty to reduce the distance between marginal probability distributions across different subjects. Tao et al. [10] mitigated statistical and semantic distribution differences by exploring diverse information across domains and incorporating metric regularization. Chen et al. [11] applied MMD to measure pairwise distribution differences, thereby achieving multi-source DA. An et al. [12] introduced STCBI-Nets, which aligns cross-brain spatiotemporal features to cut inter-subject gaps and achieves state-of-the-art results. Wu et al. [13] built a multi-source domain adaptation network that hybridizes source–target samples and aligns marginal and conditional distributions to reduce inter-subject gaps. In graph neural network-based methods, Xu et al. and Li et al. [14, 15] employed graph convolutional networks to capture the spatial and temporal relationships among different brain regions in EEG signals. Peng et

*Corresponding author

✉ 230238579@seu.edu.cn (Y. Liu); qiwei@unimap.edu.my (Q.W. Oung); hoileong@unimap.edu.my (H.L. Lee)

ORCID(s): 0000-0001-6881-9627 (Y. Liu); 0000-0001-8486-3513 (Q.W. Oung); 0000-0002-4984-2183 (H.L. Lee)

al. [16] proposed a joint feature adaptation and graph-adaptive label propagation model for EEG-based emotion recognition. Liu et al. [17] built a dynamic graph attention model that fuses multi-scale spatiotemporal cues for robust, cross-subject performance. In adversarial learning-based methods, deep transfer learning models based on Domain Adversarial Neural Networks (DANN) [18], which introduces a Gradient Reversal Layer (GRL) to optimize the classifier and discriminator components in opposing directions, have demonstrated significant potential. Li et al. [19] first applied DANN to EEG-based emotion recognition, leveraging adversarial strategies to balance domain invariance with task relevance. Pei et al. [20] extended this concept to multi-source DA. Ju et al. [21] proposed a few-shot adversarial network with multiple adversarial tasks to extract subject-invariant emotional features via label subdivision, subject mixing, and temporal scrambling. This approach was later extended with a reliable pseudo-label iteration mechanism [22], which refines pseudo-labels through perturbation-based confidence boosting and treats each emotion as a subdomain for finer alignment. In source domain selection-based methods, Guarneros et al. [23] employed central moment discrepancy as loss to measure the similarity between different source domains, thereby weighting the source domains accordingly. Zhang et al. [24] utilized weak classifiers to obtain pseudo-labels for the target domain and leveraged a small number of labels from the target domain to identify the most suitable source domain. In subspace mapping-based methods, Chai et al. [25] employed an autoencoder to map both the source and target domains into a unified space, thereby mitigating distributional differences between them. Additionally, they developed a novel linear transformation function [26] to align the marginal distributions of subspaces across different domains.

Negative transfer phenomenon refers to the adverse impact on the target domain caused by the knowledge learned by the model from the source domain. The occurrence of negative transfer is primarily due to two reasons: firstly, issues related to the quality of the source domain data, and secondly, the model's inability to identify similarities between the source and target domains. Therefore, several studies [27, 28, 29, 30, 31] focused on selecting source domain with significant differences from the target domain to reduce their impact on the model. In particular, Huang et al. [32] proposed the kernel mean matching method for estimating probability distributions, aiming to make the weighted distributions of the source and target domains as similar as possible. Tan et al. [33, 34] extended the application scenarios of instance-based transfer learning methods, introducing Transitive Transfer Learning and Distant Domain Transfer Learning, by leveraging joint matrix factorization and deep neural networks, transfer learning was applied to facilitate knowledge sharing among multiple dissimilar domains, yielding promising results. Dai et al. [35] introduced the TrAdaboost method, which effectively identifies source domain samples detrimental to transfer, and derives the supremum of the model's generalization error based on Probability Approximately Correct theory [36]. Zhang et al. [37] presented a domain transferability assessment algorithm. The existing methods focus on using a similarity metric to assess relationships between source domains. This raw data-level measurement method overlooks the fact that additional domain-related information may emerge during the training process.

The core idea of existing methods for source domain selection is to use a similarity metric to assess the differences between domains. However, this kind of approaches restricted source domain selection to the raw data-level, neglecting the changes that emerge in the data during the training process of domain adaptation models. As shown in Fig. 1, the EEG signals of a subject are chosen as the target domain, with the horizontal axis representing source domains sorted by MMD similarity metric from high to low. In the EEG-based emotion recognition task, under the labels of positive, neutral, and negative, the green-marked subjects did not achieve high accuracy corresponding to their similarity. Conversely, some low-similarity source domains obtained higher accuracy (marked in pink). The figure in the fourth row describes the relationship between accuracy and similarity, which more intuitively reflects that the correlation between the two is not so stable. Therefore, this paper focuses not only on domain differences but also on assessing whether the knowledge learned by the model in each source domain benefits the target domain, thereby re-examining the cross-subject EEG-based emotion recognition problem.

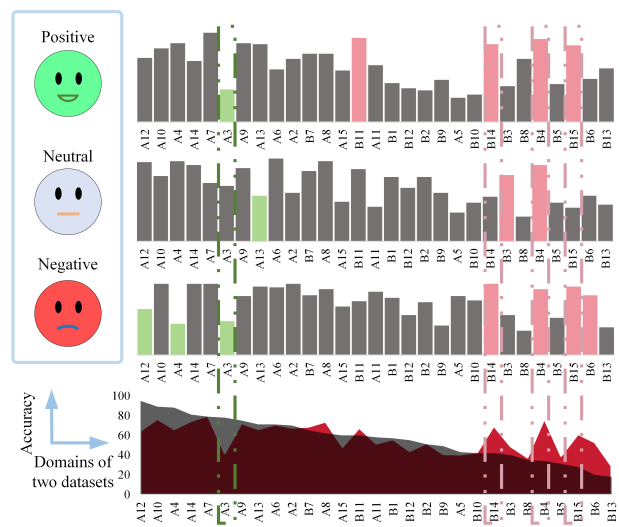

**Figure 1.** The relationship between similarity metrics of source domains and accuracy. In the horizontal axis, A and B represent different datasets.

This paper proposes a cross-subject EEG-based emotion recognition through source selection with adversarial strategy (SSAS) method, which is comprised of the source domain selection network (SS) and the domain adaptive network for adversarial strategies (AS). SSAS is capable of correctly identifying which source domain data are detrimental to knowledge transfer. By reducing the weights of these samples that contribute to negative transfer—samples that adversely affect domain adaptation training—it enhances the accuracy of cross-subject EEG-based emotion recognition. A notable aspect of SS is that it no longer focuses on seeking a similarity measure to gauge whether the knowledge learned from the source domain facilitates transfer to the target domain. The main idea of source domain selection in the SSAS method is that we cannot determine whether negative transfer phenomena manifest in the source domain during training. However, we can utilize known domain labels in the source domain to reverse-engineer the domain adaptation process, attempting to classify and determine which source domains (which are labeled) the target domain samples will be classified into. The domain classification results can reveal which source domains possess transferability (i.e., which source domains are more conducive to transfer to the target domain) after DA.

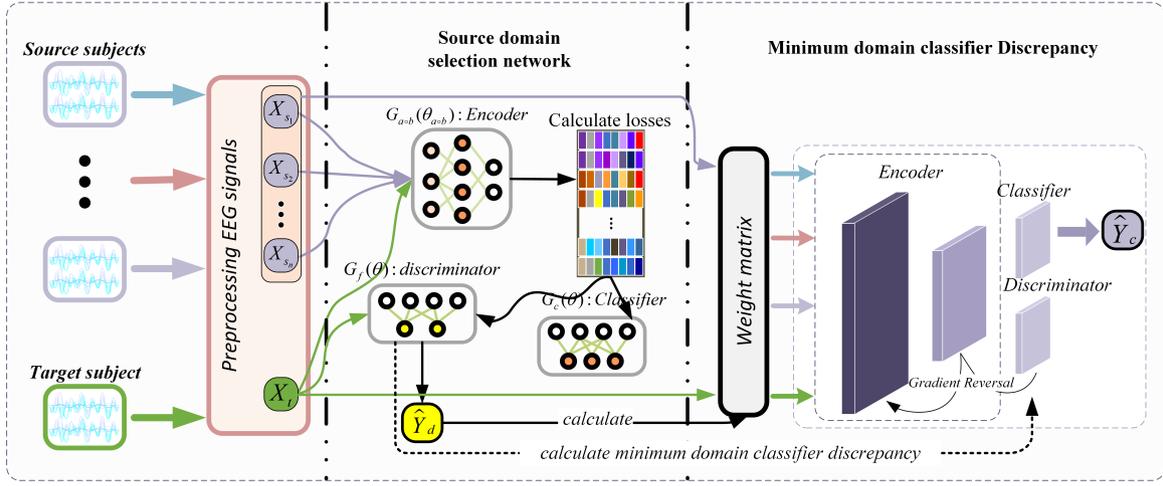

**Figure 2.** SSAS consists of two modules to achieve cross-subject EEG-based emotion recognition: the source selection network (SS) and the adversarial domain adaptation network (AS).

The rest of the paper is organized as follows. In Section 2, the proposed SSAS method is described in detail. Section 3 presents the datasets, preprocessing steps, evaluation metrics, and results, while Section 4 provides the discussion. Finally, conclusions and future work are reported in Section 5.

## 2. Proposed framework

Emotions are complex and subtle, and the classification performance in target domain highly depends on the quality of source domain. SSAS is a domain-adaptation framework with an explicit source-selection stage. This method consists of SS network and AS network.

Fig. 2 presents the SSAS framework, which comprises two main training stages: SS network and AS network. In SS, the objective is to simulate the domain adaptation process for emotion classification using domain labels, enabling the identification of source domains that better align with the target domain while avoiding negative transfer. To achieve this, the domain label prediction error is minimized alongside the reduction of distribution difference among source domains. Additionally, by adversarially maximizing the error of emotion labels, a worst-case distribution scenario is constructed for domain label classification. Ultimately, a domain discriminator F trained on the source domain is obtained to distinguish which domain the samples originate from. If certain source domains dominate the domain labels in the classification results of target domain, it indicates that the knowledge provided by these source domains in the domain adaptation is more beneficial for the target domain. AS utilizes the reorganized source domain for model training, and computes the minimum domain classifier discrepancy (MDC) using the domain discriminators F trained in SS. Gaussian noise is introduced in the feature space to enhance the model's resistance to interference, and the domain classification loss after GRL is calculated to increase the confusion level between domains. Additionally, the MMD is utilized to mitigate distributional differences among different subjects.

### 2.1. Problem definition

It is assumed that there are L labeled source domains $S_1, S_2, \dots S_L$ and an unlabeled target domain $T$. Each source domain $S_i$ is associated with two types of labels: $Y_s^c$, which represents emotion labels (e.g., positive, neutral, and negative emotions), and $Y_s^d$, which denotes domain labels (e.g., indicating the participant from whom the EEG data was collected), and it should be noted that for each $S_i$, all components of $Y_s^d$ are equal to $i$. Similarly, the target domain $T$ has the corresponding emotion labels $Y_t$. We hypothesize that all $S_i$ and $T$ lie in the same feature space and that $Y_s^c$ and $Y^t$ share the same label space. As multiple deep neural network models are employed in this paper, it is necessary to clearly define the symbols for each labeling function. We denote a deep neural network model as a labeling function $G$ with trainable weights $\theta$, where $G_\theta$ may be represented as a composite of two functions $G_c\left(G_{a \circ b}\left(\theta_{a \circ b}\right), \theta_c\right)$. Here, $G_{aob}$ represents a feature extractor network, where $a \circ b$ indicates the passage through two network layers, namely layers a and b. $G_{aob}$ takes $S_i$ and $T$ as inputs and produces an underlying feature representation denoted as $Z$, that is, $Z = G_{aob}(X; \theta_{aob})$. Meanwhile, $G_c$ is a feature labeling function to produce a classification score $\hat{y}$, i.e., $\hat{y} = G_c(Z; \theta_c)$. For clarity, all mathematical symbols are summarized in Table 1.

**Table 1**
Notation Table

| symbol | definition |
|---|---|
| $S, S_i$ | Source domain, i-th source domain |
| $T$ | Target domain |
| $\mathcal{X}$ | Instance set or input space mapping |
| $Y_c^s$ | Source domain class labels |
| $Y_d^s$ | Source domain domain labels |
| $k$ | the number of categories (emotion) |
| $\mathbb{R}^n$ | n-dimensional real vector space |
| $G_{aob}$ | Encoder |
| $G_c$ | Emotion classifier network |
| $G_d$ | Domain discriminator in the AS |
| $G_f$ | Pre-trained domain discriminator from the SS |
| $\theta, \theta_{aob}, \theta_c$ | Trainable weights |
| $f$ | Ground truth function |
| $h$ | Prediction function |
| $\epsilon_S(h)$ | Source error of $h$ |
| $\epsilon_T(h)$ | Target error of $h$ |
| $I(h)$ | Indicator function |
| $\lambda$ | Gradient reversal coefficient |
| $\lambda_e, \lambda_1', \lambda_2'$ | Minimum joint error rate of $\mathcal{H}$ |

## 2.2. Source domain selection alignment

The source domain selection alignment (SS) module is shown in Fig. 3 and the overall objective is defined by:

$$\mathcal{L}_{SS}\left(\{S_i, Y_s^c, Y_s^d\}_{i=1}^L, T, G_\theta; \theta\right) = \\ \mathcal{L}_{dcls}\left(S_i, Y_s^d, T, G_\theta; \theta\right) \\ + \alpha \sum_i^L \mathcal{L}_{mmd}\left(GRL\left(S_i, T, G_{aob}; \theta_{aob}\right)\right) \\ + \mathcal{L}_{ecls}\left(GRL\left(S_i, Y_s^c, T, G_\theta; \theta\right)\right) \quad (1)$$

where $\mathcal{L}_{dcls}$ represents the domain label classification loss, $\mathcal{L}_{ecls}$ represents the emotion label classification loss, and $\alpha$ is the balance coefficient, used to control the weight of $\mathcal{L}_{mmd}$ in the total loss.

SS does not perform emotion label classification, but rather aims to establish a model capable of identifying which domain the target domain (test set) belongs to. Due to the trade-off between discriminability and transferability in cross-subject EEG signal analysis, learning knowledge of domain-invariant representations of emotion label can lead to decreased separability between domains. Therefore, a GRL is applied to the emotion classification branch ($\mathcal{L}_{ecls}$) in the SS module, reversing the gradient of the emotion classification loss during backpropagation. This adversarial mechanism disrupts the model's learning of emotion-label information, thereby preserving (and even amplifying) the differences between domains (maintaining domain separability). $\mathcal{L}_{mmd}$ are discrepancy losses and aim to increase the distribution differences among different subjects, the ultimate purpose is to enhance the discriminability between subjects by amplifying the distributional differences among them.

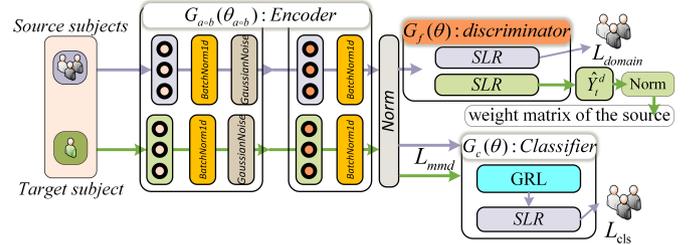

**Figure 3.** In the process of SS, class labels will be blurred and the separability of domains will be increased. This is motivated by ensuring the same domain adaptation process as AS.

As shown in Fig. 3, SS accepts preprocessed labeled and unlabeled data from T. $G_{aob}(\theta_{aob})$ is an Encoder consisting of multiple network layers. It standardizes the features of each subject by introducing batch normalization after the hidden layer. A Gaussian noise layer is utilized to enhance the model's robustness against interference. The Encoder further extracts deep features from each subject by incorporating additional hidden layers and normalization. After mapping the features to the spherical space, SS first calculates the $\mathcal{L}_{mmd}$ of the features and utilizes GRL to reverse the gradient direction during backpropagation. Subsequently, the features pass through the spherical logistic regression (SLR) layer, which serves as a classifier, to obtain domain classification results and emotion classification results. The domain classification results are obtained by $G_f(\theta)$, while the emotion classification results are obtained by $G_c(\theta)$. It is worth noting that if the model learns domain-invariant representations of emotion's label, it may lead to a decrease in the separability between domains. Therefore, GRL is also added to reverse the gradient direction during backpropagation of $\mathcal{L}_{ecls}$. After training, the domain classification results of the target domain generate a normalized weight matrix, which is used to quantify the transferability from the source domain to the target domain.

**Gradient Reversal Layer** is employed during training to enable adversarial learning between the source and target domains, thereby reducing domain differences. During the forward propagation, the GRL acts as an identity transform, leaving the input data unchanged:

$$x' = x \quad (2)$$

In the backward propagation, the GRL takes the gradient from the subsequent layers, multiplies it by $-\lambda$, and passes it to the preceding layers:

$$\frac{dG_c(\theta)}{dG_{aob}(\theta)} = -\lambda \cdot \frac{dG_c(\theta)}{dG_{aob}(\theta)} \quad (3)$$

$$\frac{d\boldsymbol{G}_f(\boldsymbol{\theta})}{d\boldsymbol{G}_{aob}(\boldsymbol{\theta})} = -\lambda \cdot \frac{d\boldsymbol{G}_f(\boldsymbol{\theta})}{d\boldsymbol{G}_{aob}(\boldsymbol{\theta})} \qquad (4)$$

Eq. (3) and Eq. (4) describe backpropagation through the GRL on the SS (affective-classification) and AS (domain-classification) branches, respectively.

**Maximum mean discrepancy** is a commonly used metric to measure the similarity between the source domain and the target domain in transfer learning. The $\mathcal{L}_{mmd}$ between the source and target domains is calculated using the following formula:

$$\mathcal{L}_{mmd} = \left\| \frac{1}{n_s} \sum_{x_i \in S_i} G_{aob}(x_i, \theta) - \frac{1}{n_t} \sum_{x_j \in T} G_{aob}(x_j, \theta) \right\|_F^2 \qquad (5)$$

where $n_s$ and $n_t$ represent the sample sizes of the source and target domains respectively, and $G_{aob}(x_i, \theta)$ represents the features extracted from the EEG data by the Encoder.

**Gaussian noise.** Introducing Gaussian Noise can make the model less sensitive to minor perturbations in the input data, thus enhancing its robustness [38]. Meanwhile, certain emotional states may exhibit significant uncertainty, which Gaussian Noise can simulate. By simulating more realistic fluctuations in emotional signal, the model can better adapt to various forms of emotional expression. For these reasons, a noised feature vector $\delta \in \mathbb{R}^n$ is added to each learned feature sample $Z$ during training in both the SS and AS modules, as formulated in Eq. (6). This noise is only used during the training phase to promote robustness and does not affect the final classification branch. During inference, noise is disabled, ensuring that model performance remains unaffected.

$$\begin{cases} \delta \sim N(0, v) \\ Z = Z + \delta \end{cases} \qquad (6)$$

where $\delta$ is drawn from a normal distribution with zero-mean and variance $v$. Both $\delta$ and the feature vector $Z \in \mathbb{R}^n$ share the same dimensionality.

**Spherical Logistic Regression (SLR)** [39] is employed to compute the predictive scores for each category of deep features $Z$ on the sphere $\mathbb{Q}_r^{n-1}$. Here, $\mathbb{Q}_r^{n-1} = \{Z \in \mathbb{R}^n \mid \|Z\|_2 = r\}$ denotes an $(n-1)$-dimensional hypersphere of radius $r$ in $\mathbb{R}^n$, where $r$ defines the radius of the feature space on which the deep features are normalized. Circles on the sphere are defined as $\omega^T Z + b = 0$, where $\omega$ is a unit normal vector and $b \in [-r, r]$. The definition of the SLR layer is as follows:

$$p(y = k \mid Z) \propto \exp\left(\omega_k^T Z + b_k\right), \quad k = 1, 2, \ldots, K \qquad (7)$$

where $\omega_k \in \mathbb{R}^n$ with $\|\omega_k\| = 1$, and $b_k$ is a learnable bias parameter.

**Weight matrix of the source domain** depends on the model's classification results for the target domain, where the number of categories k is determined by the quantity of source domains $S_i$. SS outputs the classification results $\widehat{Y}_s^d \in \mathbb{R}^N$, where $N$ is the number of source domain samples. The importance of each source domain relative to the target domain is measured by computing the proportion of domain numbers in $\widehat{Y}_s^d$. The formula for the weight matrix, denoted as $weight$, is as follows:

$$weight_l = \min + \frac{(\max - \min) \cdot \left(count_l - count_{\min}\right)}{count_{\max} - count_{\min}} \qquad (8)$$

where min and max represent the normalized range, which in this paper is set to 0.5 and 2. The count stores the sample quantities for each category (domain label); $count \in R^L$, where the elements L represent the number of source domains. The minimum and maximum values in count are denoted as $count_{\min}$ and $count_{\max}$, respectively.

## 2.3. Domain adaptive network for adversarial strategies

The domain adaptive network for adversarial strategies (AS) is shown in Fig. 4 and the overall objective is defined as follows:

$$\begin{aligned} \mathcal{L}_{AS}\left(\left\{\boldsymbol{S}_i, \boldsymbol{Y}_s^c, \boldsymbol{Y}_s^d\right\}_{i=1}^L, \boldsymbol{T}, \boldsymbol{G}_\theta; \boldsymbol{\theta}\right) = \\ \mathcal{L}_{dcls}\left(\text{GRL}\left(\boldsymbol{S}_i, \boldsymbol{Y}_s^d, \boldsymbol{G}_d; \boldsymbol{\theta}\right)\right) \\ + \mathcal{L}_{ecls}\left(\boldsymbol{S}_i, \boldsymbol{Y}_s^c, \boldsymbol{T}, \boldsymbol{G}_c; \boldsymbol{\theta}\right) \\ + \alpha \cdot \sum_i^L \mathcal{L}_{mmd}\left(\boldsymbol{S}_i, \boldsymbol{Y}_s^c, \boldsymbol{T}, \boldsymbol{G}_{aob}; \boldsymbol{\theta}\right) \\ + \mathcal{L}_{mdc}\left(\boldsymbol{S}_i, \boldsymbol{Y}_s^d, \boldsymbol{G}_f, \boldsymbol{G}_d; \boldsymbol{\theta}\right) \end{aligned} \qquad (9)$$

The difference between AS and SS lies in the application of the GRL and the minimum domain classifier discrepancy loss ($\mathcal{L}_{mdc}$). First, the role of GRL in AS is the opposite of its role in SS. In AS, a GRL is attached to the domain classification branch ($\mathcal{L}_{dcls}$), so the gradient of the domain classification loss is reversed during backpropagation. This adversarial setup confuses the domain discriminator and forces the feature extractor to learn domain-invariant features. In contrast, the emotion classification branch ($\mathcal{L}_{ecls}$) in AS is trained normally without a GRL (i.e. its loss is backpropagated without negation). This design ensures that the source and target feature distributions become more similar, allowing the model to better adapt to the target domain. Placing the GRL on different branches leads to different objectives: with SS, a GRL on the emotion branch hinders emotion-label learning to preserve domain separability; with AS, a GRL on the domain branch hinders domain-label learning to enforce domain invariance. Second, $\mathcal{L}_{mdc}$ helps balance the adversarial training so that the domain discriminator $G_d$ does not lose its ability to distinguish domains too early.

As shown in Fig. 4, the AS method receives the weight matrix and the high-performing domain discriminators $\boldsymbol{G}_f$. The source domain data is reconstructed according to the weight matrix, and the reconstructed data is passed through $\boldsymbol{G}_{aob}(\boldsymbol{\theta}_{aob})$ to generate deep features $Z$. As shown in Eq. (9), $\widehat{Y}_s^d$ output by $\boldsymbol{G}_d$ is used to calculate the cross-entropy loss with $Y_s^d$; The MDC loss is computed based on the two types of predicted domain labels, $\hat{Y}_s^d$, output by $\boldsymbol{G}_f$ and $\boldsymbol{G}_d$, respectively. the predicted emotion labels output by $\boldsymbol{G}_c$ are used to compute the cross-entropy loss to minimize the classification error rate on the source domain; pseudo-labels of the target domain are also output by $\boldsymbol{G}_c$ and used for calculating the MMD loss.

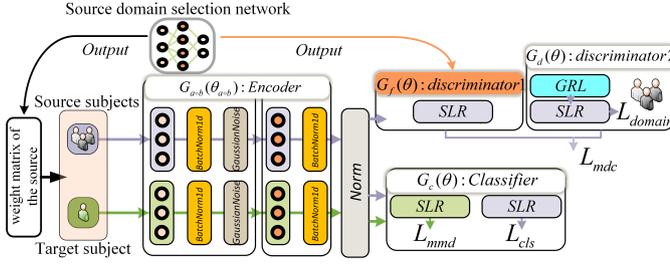

**Figure 4.** AS enhances the model's generalization ability through two aspects: domain separability and distribution discrepancy.

**Minimum domain classifier discrepancy loss.** In this paper, the absolute difference of the probability outputs from two domain discriminators is utilized as the discrepancy loss, as shown in the following equation:

$$\mathcal{L}_{mdc} = d\left(p_d, p_f\right) = \frac{1}{L} \sum_{i=1}^{L} \left| p_d - p_f \right| \qquad (10)$$

where $p_d$ and $p_f$ represent the probability output by the domain discriminators $\boldsymbol{G}_d$ and $\boldsymbol{G}_f$, respectively.

The design purpose of $\mathcal{L}_{mdc}$ is to balance the adversarial relationship between domain labels and class labels, rather than allowing the domain discriminators $\boldsymbol{G}_d$ to prematurely lose its domain separability. The benefit of doing so lies in enabling the formation of a new adversarial relationship between the domain discriminator $\boldsymbol{G}_d$ and $\boldsymbol{G}_f$, which allows the model to adapt to a wider range of distributions during training, thereby achieving better performance in the target domain.

## 2.4. Theoretical insights of SSAS

SS doesn't adhere to the conventional approach of source domain selection. Instead, it uses domain labels to reverse-engineer the DA process for class labels. The theoretical foundation for SSAS is inspired by the work of Ben-David et al. [40] and Lu et al. [41], which discuss bounds on expected error in target domains for domain adaptation and generalization. In this subsection, we aim to illustrate the relationship between SSAS and these two theories.

In the theory by Ben-David et al., the expected error on the target domain $\varepsilon_T(h)$ is bounded by three key terms: (i) expected error on the source domain, $\varepsilon_S(h)$; (ii) $\mathcal{H}\Delta\mathcal{H}$-distance ($d_{\mathcal{H}\Delta\mathcal{H}}(S,T)$), which is measured as the discrepancy between two classifiers; and (iii) the shared error of the ideal joint hypothesis, $\lambda$. Lu et al. introduced a theoretical framework that bounds the error on the unseen target domain, which introduced $\mathcal{H}$-distance ($d_{\mathcal{H}}(S,T)$) for domain divergence. The two theories and their relationships can be explained as follows.

Let $\mathcal{X}$ be the input space and $\mathcal{H}$ be a class of hypotheses corresponding to this space, for any $h \in \mathcal{H}$, the following inequality holds with at least a probability of $1 - \delta$:

$$\varepsilon_T(h) \leq \varepsilon_S(h) + \frac{1}{2} d_{\mathcal{H}\Delta\mathcal{H}}(S,T) + \lambda \qquad (11)$$

where $1 - \delta$ represents the confidence level of the inequality.

For a hypothesis space $\mathcal{H}$, its corresponding $\mathcal{H}\Delta\mathcal{H}$ space is:

$$g \in \mathcal{H}\Delta\mathcal{H} \Leftrightarrow g(x) = h(x) \oplus h'(x) \text{ for some } h, h' \in \mathcal{H} \qquad (12)$$

The symbol "$\oplus$" denotes the exclusive OR. Inspired by the domain generalization theory proposed by Lu et al., which constrains the expected error on target samples, the domain adaptation bounds for expected error on target samples were reconstructed in this paper.

**Proposition 1** Given two probability distributions $S_i$ and $T$, defined on the input space $\mathcal{X}$, representing the probability distributions of the source domain and the target domain, respectively. $\mathcal{H}$ denotes the hypothesis space on $\mathcal{X}$, and the probability distribution of multi-source domains can be expressed as $S$, and let $\varphi_i$ be a collection of non-negative coefficient with $\sum_i^L \varphi_i = 1$. Let $\mathbb{O}$ be a set of distribution s.t. $\forall S \in \mathbb{O}$, the following holds

$$d_{\mathcal{H}\Delta\mathcal{H}}\left(\sum_i^L \varphi_i S_i, S\right) \leq \max_{j,k} d_{\mathcal{H}\Delta\mathcal{H}}\left(S_j, S_k\right) \qquad (13)$$

Then, for any $h \in \mathcal{H}$,

$$\varepsilon_T(h) \leq \sum_i^L \varphi_i \varepsilon_{S_i}(h) + \frac{1}{2} \min_{S \in \mathbb{O}} d_{\mathcal{H}\Delta\mathcal{H}}(S,T)$$
$$+ \frac{1}{2} \max_{j,k} d_{\mathcal{H}\Delta\mathcal{H}}\left(S_j, S_k\right) + \lambda_e \qquad (14)$$

Here, $\lambda_e$ represents the minimum joint error rate of $\mathcal{H}$ on $S_i$ and $T$; $\varepsilon_{S_i}(h)$ denotes the error rate of hypothesis $h$ mapping from $\mathcal{X}$ to the classification set on distribution $S_i$. $\min_{S \in \mathbb{O}} d_{\mathcal{H}\Delta\mathcal{H}}(S,T)$ represents the $\mathcal{H}\Delta\mathcal{H}$-divergence of $S$ and $T$ under the hypothesis space $\mathcal{H} \triangle \mathcal{H}$, indicating the distribution discrepancy between different domains. Eq. (14) expands the target domain error bound by considering worst-case distributional scenarios, which capture additional variations in domain shifts and thereby introduce a new form of constraint. This approach enhances the model's adaptability across a broader range of transfer scenarios. The design motivation for SSAS originates from the right side of Eq. (14), where SSAS assumes that the expected error on the target domain is determined by these four terms on the right. The first term in Eq. (14), $\sum_i^L \varphi_i \varepsilon_{S_i}(h)$ exists in almost all methods and can be minimized via supervision from class labels with cross-entropy loss. This term corresponds to $\mathcal{L}_{dcls}\left(\boldsymbol{S}_i, \boldsymbol{Y}_s^d, \boldsymbol{T}, \boldsymbol{G_\theta}; \theta\right)$ and $\mathcal{L}_{ecls}\left(\boldsymbol{S}_i, \boldsymbol{Y}_s^c, \boldsymbol{T}, \boldsymbol{G_\theta}; \theta\right)$ in SS and AS, respectively. The second term, $\min_{S \in \mathbb{O}} d_{\mathcal{H}\Delta\mathcal{H}}(S,T)$, is the core step of DA, explaining why MMD is used to minimize the distribution discrepancy between different domains in both AS and SS. As the goal of SSAS is to find a model that performs well on the target domain, we need a metric to minimize the value of $\min_{S \in \mathbb{O}} d_{H \Delta H}(S, T)$ to reduce $\varepsilon_T(h)$. The third term, $\max_{j,k} d_{H \Delta H}(S_j, S_k)$, does not directly reduce the Supremum of $\varepsilon_T(h)$, but by maximizing this term, it enlarges the range of $\mathbb{O}$ in the second term, $\min_{S \in \mathbb{O}} d_{H \Delta H}(S, T)$, thereby reducing the infimum of $\min_{S \in \mathbb{O}} d_{H \Delta H}(S, T)$. This term corresponds to $\mathcal{L}_{ecls} \left( GRL \left( S_i, Y^c_s, T, G_\theta; \theta \right) \right)$ and $\mathcal{L}_{mdc} \left( S_i, Y^d_s, G_f, G_d; \theta \right)$ in SS and AS, respectively. In SS, the attempt is made to provide a "worst-case" distribution scenario for domain discrimination by maximizing the cross-entropy loss of class labels, aiming to increase the range of $\mathbb{O}$ in the third term, $\min_{S \in \mathbb{O}} d_{H \Delta H}(S, T)$. The motivation behind designing the $\mathcal{L}_{mdc}$ in AS comes from $\max_{j,k} d_{H \triangle H}(S_j, S_k)$. Similar to other DA methods, AS uses GRL to maximize the cross-entropy loss with the domain labels; however, we additionally introduce $\mathcal{L}_{mdc} \left( S_i, Y^d_s, G_f, G_d; \theta \right)$ to achieve the function of $\max_{j,k} d_{H \Delta H}(S_j, S_k)$. $\mathcal{L}_{mdc}$ induces an adversarial relationship between $G_f$ and $G_d$, thereby avoiding premature blurring of the separability of domains and reducing the infimum of $\min_{S \in \mathbb{O}} d_{H \triangle H}(S, T)$.

**Table 2**
Setting values for hyperparameters in SSAS.

| Hyperparameter | SEED / SEED-IV | HBUED |
|---|---|---|
| batch size | 50 | 50 |
| learning rate | 0.01 | 0.001 |
| epochs of SS | 20 | 20 |
| epochs of AS | 30 | 40 |
| $\lambda$ | 1 | 1 |
| $\alpha$ | 0.5 | 0.5 |
| dimension after $G_{aob}$ | 128 | 32 |
| dimension after $G_c$ | 3/4 | 2 |
| dimension after $G_d$ | 14 | 49 |

The concept in SS is to utilize domain labels to reverse-engineer the DA process for classification tasks. Whether in AS or SS, the error rate of the hypothesis $h$ mapping from $\mathcal{X}$ to the classification set in the target domain is determined by Eq. (14). This reflects the similarity in the training process between AS and SS, allowing SS to quantify the transferability of the source domain when the model performs class labels tasks, thereby selecting out data that is useful for the target domain for model training.

## 3. EXPERIMENTS

### 3.1. Experimental setup

Three publicly available EEG emotion datasets are utilized in this paper, SEED, SEED-IV and HBUED, to validate the effectiveness of the SSAS method.

**SEED** [43] is designed for EEG-based emotion classification tasks. It records EEG signals from 15 subjects while they watch positive, neutral, and negative emotional movie clips, with each video segment lasting $3 - 5$ minutes. Each subject underwent three days of data collection, with approximately one week between sessions. The EEG signals were recorded using a 62-channel ESI NeuroScan device at a sampling rate of 1 kHz, then downsampled to 200 Hz, and band-pass filtered from 0 to 75 Hz to remove artifacts from the EEG signal.

**SEED-IV.** [44] Data from 15 subjects (7 males, 8 females) were collected using a 64-channel ESI NeuroScan system and an SMI eye tracker for the SEED-IV dataset. Unlike the SEED dataset, each subject in the SEED-IV dataset was required to watch four different emotion-inducing movie clips during each experimental session. Each emotion category consisted of 6 segments, totaling 6 segments each for 'happiness,' 'sadness,' 'fear,' and 'neutral' movie clips. EEG data were downsampled to 200 Hz and band-pass filtered within the frequency range of 1-70 Hz.

Both the SEED and SEED-IV datasets provide precomputed differential entropy (DE) features, with details on the feature extraction methods available in [43, 44]. DE features are extracted from each segment at five frequency bands: delta (1–4 Hz), theta (4–8 Hz), alpha (8–14 Hz), beta (14–31 Hz), and gamma (31–50 Hz). In addition, the EEG features are smoothed to minimize the artifacts in the EEG features using the linear dynamic system approach [45].

**HBUED** [46] data from 50 subjects (24 males, 26 females). This dataset employs video stimuli for emotion induction, comprising 24 clips covering all four quadrants of the VA model—six clips per category, each lasting 2 min. The EEG data were recorded using a 32-channel Brain Products device following the International 10–20 system at a sampling rate of 1000 Hz. Preprocessing included: 49–51 Hz notch filtering to eliminate power line interference; 0.1–48 Hz bandpass filtering; downsampling to 128 Hz; 3-second baseline correction; and ICA artifact removal. SSAS utilized the officially provided DE features as initial inputs, following the paper.

### 3.2. Implementation details

In both SS and AS methods, the stochastic gradient descent (SGD) optimizer [47] was employed. The SS method was trained for 20 epochs, while the AS method was trained for 30 epochs. Beyond these training epochs, the classification results of the model didn't change significantly. To ensure model consistency, a batch size of 50 was set for both datasets, momentum was set to 0.9 and weight factor was set to 0.005. The hyperparameter $\lambda$ for the GRL in both SS and AS was set to 1. Detailed parameter settings are shown in Table 2.

Python 3.11.4 and PyTorch 2.0.1 were utilized to implement the SSAS method. All experimental results were obtained on a computer equipped with an Intel (R) Core (TM) $i7 - 8700$ CPU and an NVIDIA GeForce RTX 4090 GPU.

Both SEED and SEED-IV consist of three sessions from 15 subjects, while HBUED contains data from 50 subjects. For all datasets, we employed Leave-One-Subject-Out Cross-Validation (LOSOCV), where in each experiment, one subject was selected as the target domain, and the remaining

**Table 3**
Accuracy performance ( % ) comparison with different studies that performed LOSOCV on SEED and SEED-IV datasets.

| Methods | Evaluation Index | | | | | | |
|---|---|---|---|---|---|---|---|
| | Best session | Session 1 | Session 2 | Session 3 | Average | F1-score | AUC |
| **SEED** | | | | | | | |
| Nontransfer | 67.99 ± 10.64 | 56.73 ± 16.29 | 67.99 ± 10.64 | 67.03 ± 10.05 | 63.92 | 69.22 ± 08.74 | 72.50 ± 09.03 |
| TCA [6] | 75.32 ± 10.84 | 71.80 ± 13.99 | 67.99 ± 10.64 | 75.32 ± 10.84 | 66.68 | 75.17 ± 11.16 | 81.42 ± 08.42 |
| DAN [19] | 66.51 ± 08.14 | 64.43 ± 05.91 | 65.06 ± 08.14 | 66.51 ± 08.14 | 65.34 | — | — |
| MAS-DGAT-Net [17] | 80.02 ± 05.79 | 80.02 ± 05.79 | — | — | — | — | — |
| SAAE [25] | 80.34 ± 06.13 | 80.34 ± 06.13 | 74.68 ± 12.76 | 78.73 ± 12.96 | 77.92 | — | — |
| ASFM [26] | 83.51 ± 07.40 | 83.51 ± 07.40 | 76.68 ± 12.16 | 81.20 ± 10.68 | 80.46 | — | — |
| WGAN [9] | 87.07 ± 07.14 | 87.07 ± 07.14 | — | — | — | — | — |
| MFSAN [8] | 81.19 ± 09.29 | 75.70 ± 10.86 | 81.19 ± 09.29 | 79.19 ± 08.17 | 78.69 | — | — |
| MADA [20] | 82.52 ± 07.02 | 78.67 ± 08.26 | 82.52 ± 07.02 | 80.05 ± 8.40 | 80.41 | — | — |
| RGNN [38] | 85.30 ± 06.72 | 85.30 ± 06.72 | — | — | — | — | — |
| MACI [10] | 77.43 ± 07.37 | 73.42 ± 06.86 | 70.81 ± 06.18 | 77.43 ± 07.37 | 73.89 | — | — |
| MS-MDA [11] | 82.88 ± 07.83 | 82.88 ± 07.83 | 80.93 ± 10.16 | 76.88 ± 14.16 | 80.23 | — | — |
| MWACN [7] | 89.30 ± 09.18 | 89.30 ± 09.18 | — | — | — | — | — |
| GRU-Conv [14] | 87.04 ± 13.35 | 87.04 ± 13.35 | — | — | — | — | — |
| MFA-LR [23] | 89.11 ± 07.72 | 89.11 ± 07.72 | 82.07 ± 12.38 | 84.64 ± 12.42 | 85.27 | — | — |
| MGFKD [24] | 87.51 ± 07.68 | 87.51 ± 07.68 | — | — | — | — | — |
| STCBI-Nets [12] | 90.21 ± 06.32 | 90.21 ± 06.32 | — | — | — | — | — |
| SH-MDA [13] | 90.27 ± 05.56 | 90.27 ± 05.56 | — | — | — | — | — |
| EEGMatch [42] | 91.35 ± 07.03 | 91.35 ± 07.03 | — | — | — | — | — |
| Gusa [15] | 91.77 ± 05.91 | 91.77 ± 05.91 | — | — | — | — | — |
| **SSAS(Ours)** | **91.97 ± 07.82** | **91.97 ± 07.82** | **83.56 ± 13.74** | **86.43 ± 11.18** | **87.32** | 90.23 ± 09.25 | 92.68 ± 06.92 |
| **SEED-IV** | | | | | | | |
| Nontransfer | 44.59 ± 13.69 | 44.59 ± 13.69 | 37.99 ± 12.56 | 38.99 ± 12.36 | 40.52 | 42.55 ± 12.49 | 63.38 ± 12.28 |
| TCA [6] | 58.03 ± 13.14 | 58.03 ± 13.14 | 56.56 ± 13.77 | 57.16 ± 14.43 | 57.25 | 56.75 ± 14.12 | 71.89 ± 09.09 |
| DAN [19] | 41.27 ± 07.98 | 35.73 ± 06.35 | 38.28 ± 07.28 | 41.27 ± 07.98 | 38.42 | — | — |
| RGNN [38] | 57.37 ± 08.81 | 53.49 ± 10.57 | 57.37 ± 08.81 | 53.97 ± 16.93 | 54.94 | — | — |
| MS-MDA [11] | 67.25 ± 05.24 | 67.25 ± 05.24 | 62.73 ± 09.37 | 63.71 ± 08.39 | 64.56 | — | — |
| JAGP [16] | 60.29 ± 15.14 | 54.37 ± 09.49 | 51.06 ± 15.14 | 60.29 ± 15.14 | 55.24 | — | — |
| MFA-LR [23] | 74.99 ± 12.10 | 64.96 ± 14.04 | 74.99 ± 12.10 | 68.78 ± 16.16 | 69.58 | — | — |
| MGFKD [24] | 67.80 ± 08.25 | — | 67.80 ± 08.25 | — | — | — | — |
| EEGMatch [42] | 65.53 ± 08.31 | 65.53 ± 08.31 | — | — | — | — | — |
| MAS-DGAT-Net [17] | 70.22 ± 09.12 | — | 70.22 ± 09.12 | — | — | — | — |
| SH-MDA [13] | 73.41 ± 08.27 | — | 73.41 ± 08.27 | — | — | — | — |
| Gusa [15] | 75.17 ± 08.82 | — | 75.17 ± 08.82 | — | — | — | — |
| STCBI-Nets [12] | 75.12 ± 07.99 | — | 75.12 ± 07.99 | — | — | — | — |
| **SSAS(Ours)** | **77.99 ± 10.56** | **68.43 ± 12.58** | **77.99 ± 10.56** | **73.85 ± 14.10** | **73.42** | 74.33 ± 10.91 | 83.33 ± 07.28 |

¯ indicates that the experiment results are not reported.

subjects were used as source domains. This results in 45 experiments for SEED and SEED-IV, and 50 experiments for HBUED.

## 3.3. Experimental results

To validate the advantages of SSAS on the SEED, SEED-IV and HBUED datasets, we present experimental results of various existing methods in Table 3 and Table 4. The hyperparameter settings for the methods listed in the table remain consistent, except for those directly obtained from the original papers. It is noteworthy that both the Nontransfer method, which trains directly on cross-subject data without transfer learning, and the TCA method utilize the same DE features as inputs as SSAS for a more intuitive comparison. The "-" in the table indicates that results were not provided in the original papers, and the source code for replication experiments was not available.

**Table 4**
Accuracy ( % ) comparison with different studies that performed LOSOCV on HBUED datasets.

| Method | Valence | Arousal | Four-class |
|--------|---------|---------|------------|
| Nontrasfer | 54.34 / 06.15 | 52.47 / 04.22 | 29.11 / 09.13 |
| TCA [6] | 61.54 / 05.19 | 60.99 / 05.86 | 31.81 / 05.95 |
| DAN [19] | 65.34 / 06.15 | 66.47 / 04.22 | 54.11 / 09.13 |
| SAAE [25] | 69.54 / 05.19 | 68.79 / 05.86 | 55.81 / 05.95 |
| MFA-LR [23] | 73.13 / 08.46 | 72.84 / 08.16 | 59.75 / 08.96 |
| SSAS(*Ours*) | **76.34 / 10.34** | **75.57 / 09.53** | **61.62 / 08.69** |

The results show that SSAS significantly outperforms the comparison methods. On the SEED dataset, SSAS achieved the highest accuracy in Session 1 (91.97%), surpassing 20 existing methods. Compared to Nontransfer, SSAS improved accuracy by approximately 33.52% in the best session. Moreover, SSAS outperformed state-of-the-art (SOTA) methods across all three sessions, with an average improvement of at least 2.05%. On the SEED-IV dataset, SSAS achieved the highest accuracy in Session 2 (77.99%) among 12 existing methods, demonstrating an approximate 33.4% increase over Nontransfer. In Sessions 1 and 3, SSAS also outperformed all SOTA methods, with improvements of 1.18% and 5.07%.

Additionally, Table 3 provides detailed AUC and F1 scores for SSAS on the SEED and SEED-IV datasets. Regarding precision and recall, SSAS achieved average F1 scores of 90.23 and 74.33 across the three sessions on SEED and SEED-IV, respectively. In terms of classification performance, SSAS obtained average AUC scores of 92.68 and 83.33 for the SEED and SEED-IV datasets. On the HBUED dataset, SSAS also achieved substantial improvements, outperforming all baseline methods in valence, arousal, and four-class classification tasks, as shown in Table 4. Table 5 further presents SSAS's performance for each subject in the SEED and SEED-IV datasets.

### 3.4. Confusion matrix

Fig. 5 depicts the confusion matrices of SSAS in SEED and SEED-IV, representing the accuracy for each emotion. The three confusion matrices in the first column correspond to the three session segments of the SEED dataset, while the second column represents the three session segments of SEED-IV.

For SEED, it can be observed that negative emotions are more easily recognized, while confusion occurs between neutral and positive emotions. This finding suggests that negative emotions are more easily conveyed among subjects. Regarding SEED-IV, neutral emotions are relatively easier to distinguish, while confusion is more likely to occur between sad, fearful, and happy. These results indicate that when subjects experience neutral emotions, they are the most distinguishable, hence necessitating further research

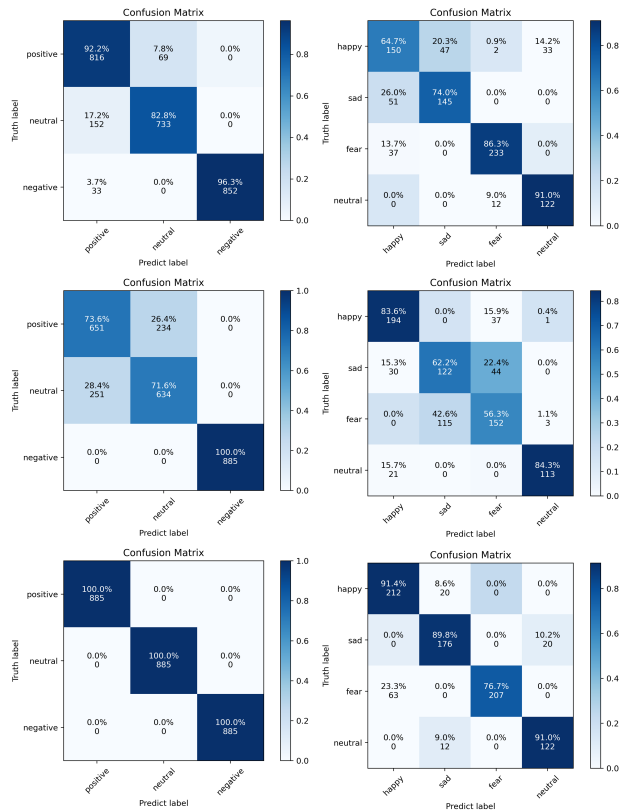

**Figure 5.** The confusion matrices of SSAS on SEED and SEED-IV datasets.

on enhancing the transferability and separability between contrasting emotions.

## 4. DISCUSSION

### 4.1. Visualization

For better understanding of the effectiveness of SSAS, 50 EEG samples were randomly selected from each subject in the SEED and SEED-IV for visualization using t-SNE [48]. As shown in Fig. 6, the first column corresponds to the SEED, while the second column corresponds to the SEED-IV. Different shapes indicate different categories, and different colors represent different subjects. It is evident that these two datasets exhibit disordered raw signals. After feature extraction with SSAS, the disorderliness of the distribution of each subject's data significantly decreases. As shown in the second row of Fig. 6, the distribution of the target domain tends to resemble that of the source domains with good transferability, and the overall data distribution demonstrates orderliness. Furthermore, for all subjects after SS processing, samples of the same emotion category become closer, while samples of different categories become farther apart (as observed in each cluster). The third row of Fig. 6 illustrates the t-SNE after the complete SSAS stage. It is apparent that the separability of classes for all participants is further enhanced, with the sample distribution forming distinct three or four clusters, corresponding to the number of emotion labels. These observations suggest that SSAS not

**Table 5**

Accuracy performance ( % ) and standard deviation of SSAS on SEED and SEED-IV datasets for each available session, using LOSOCV.

| | **SEED** | | | **SEED-IV** | | |
|---|---|---|---|---|---|---|
| | *Session* 1 | *Session* 2 | *Session* 3 | *Session* 1 | *Session* 2 | *Session* 3 |
| *Subject* 1 | 100 ± 03.97 | 98.98 ± 03.66 | 78.72 ± 00.82 | 69.92 ± 03.16 | 73.32 ± 02.10 | 80.29 ± 03.13 |
| *Subject* 2 | 80.79 ± 00.64 | 84.60 ± 01.58 | 69.68 ± 01.74 | 82.37 ± 10.56 | 95.55 ± 04.64 | 87.10 ± 01.73 |
| *Subject* 3 | 88.59 ± 02.68 | 100 ± 00.00 | 93.33 ± 01.07 | 74.15 ± 02.67 | 91.95 ± 05.51 | 82.85 ± 04.85 |
| *Subject* 4 | 99.25 ± 00.67 | 87.27 ± 02.28 | 85.08 ± 01.15 | 60.99 ± 02.26 | 72.60 ± 03.82 | 84.55 ± 01.78 |
| *Subject* 5 | 89.64 ± 01.61 | 100 ± 00.00 | 92.62 ± 01.41 | 60.63 ± 09.37 | 79.45 ± 01.36 | 91.24 ± 04.57 |
| *Subject* 6 | 100 ± 00.00 | 75.93 ± 01.99 | 98.79 ± 02.02 | 35.02 ± 02.79 | 79.45 ± 05.36 | 64.48 ± 02.06 |
| *Subject* 7 | 86.59 ± 01.52 | 80.49 ± 02.36 | 85.80 ± 02.00 | 74.74 ± 03.70 | 82.21 ± 02.09 | 89.78 ± 02.77 |
| *Subject* 8 | 77.21 ± 01.23 | 78.19 ± 00.94 | 93.56 ± 01.37 | 75.44 ± 04.08 | 91.11 ± 05.59 | 74.21 ± 01.82 |
| *Subject* 9 | 98.53 ± 03.72 | 100 ± 00.00 | 94.01 ± 02.20 | 74.97 ± 03.42 | 78.25 ± 04.39 | 38.93 ± 03.98 |
| *Subject* 10 | 86.67 ± 01.66 | 65.54 ± 02.44 | 74.58 ± 00.85 | 81.55 ± 02.43 | 71.75 ± 04.24 | 69.95 ± 02.98 |
| *Subject* 11 | 99.06 ± 03.33 | 81.24 ± 03.61 | 85.20 ± 01.60 | 75.79 ± 04.23 | 59.38 ± 01.99 | 64.60 ± 02.68 |
| *Subject* 12 | 99.47 ± 01.83 | 71.64 ± 00.99 | 61.09 ± 01.92 | 52.17 ± 03.34 | 60.82 ± 02.38 | 57.79 ± 02.32 |
| *Subject* 13 | 81.73 ± 01.83 | 55.22 ± 01.35 | 87.98 ± 02.51 | 76.62 ± 04.15 | 78.12 ± 02.52 | 70.68 ± 02.88 |
| *Subject* 14 | 92.02 ± 01.22 | 77.55 ± 03.59 | 96.61 ± 03.70 | 71.21 ± 05.07 | 69.83 ± 01.80 | 83.82 ± 04.86 |
| *Subject* 15 | 100 ± 01.59 | 96.72 ± 02.61 | 99.44 ± 02.14 | 60.87 ± 06.29 | 86.18 ± 04.35 | 67.52 ± 02.26 |

only reduces the distribution discrepancy of EEG signals between the source and target domains but also improves the separation of feature samples related to different emotion categories.

## 4.2. The impact of iteration times on SS

Fig. 7 visualizes the evolution of source-domain weights over iterations in the SS module, where higher weights are shown in yellow and lower weights in blue. The images in the first and second rows respectively depict the top five subjects of the SEED dataset and the SEED-IV dataset. For the SEED dataset, over the course of 20 iterations, the weights for certain source domains steadily increase. This indicates that the SS module is progressively identifying and emphasizing more transferable source domains (those contributing positively to the target domain) as training iterates. In contrast, the SEED-IV dataset does not exhibit a clear monotonic increase in weights. Instead, several source domains in SEED-IV remain low-weight (blue) throughout the iterations, signaling that these domains have persistently poor transferability (i.e., they contribute to negative transfer and make cross-subject adaptation more difficult).

These observations suggest that overall SEED-IV has weaker cross-domain transferability, which in turn makes accuracy improvements more challenging on that dataset. Notably, the SS module flags a larger number of source subjects as unhelpful (low-weight) in SEED-IV than in SEED. By down-weighting these less transferable sources, SSAS effectively focuses on the subset of source domains that are truly beneficial. Consequently, the performance gain from source selection is more pronounced on SEED-IV. This is evidenced by our results – for example, removing the SS component causes a larger drop in accuracy on SEED-IV than on SEED (as shown by a 7% vs 5% decrease in Table 7) – highlighting that filtering out detrimental sources yields greater benefits in the more heterogeneous SEED-IV scenario. In summary, the iterative SS process proves especially valuable when source data quality or relevance is highly variable, as it can isolate and mitigate negative transfer from poorly suited source domains.

However, in scenarios where all source domains are of uniformly high quality and negative transfer is negligible (as in the SEED dataset), the advantage of using the SS module is naturally limited. In such cases, our current source selection strategy provides only marginal accuracy gains that may not justify the additional training complexity. This suggests that when inter-source variability is low, the present SS approach yields diminishing returns. As a future direction, we propose learning more fine-grained source weights or integrating a lightweight selection mechanism to better handle situations with minimal distribution shift, ensuring the source selection process remains beneficial even when domain quality is uniformly high.

## 4.3. Effect of the number of source domains

We evaluate robustness by varying the number of source domains $S \in \{2, 5, 10, 15\}$ under the same LOSOCV protocol, comparing (i) randomly sampled sources with (ii) Top-S

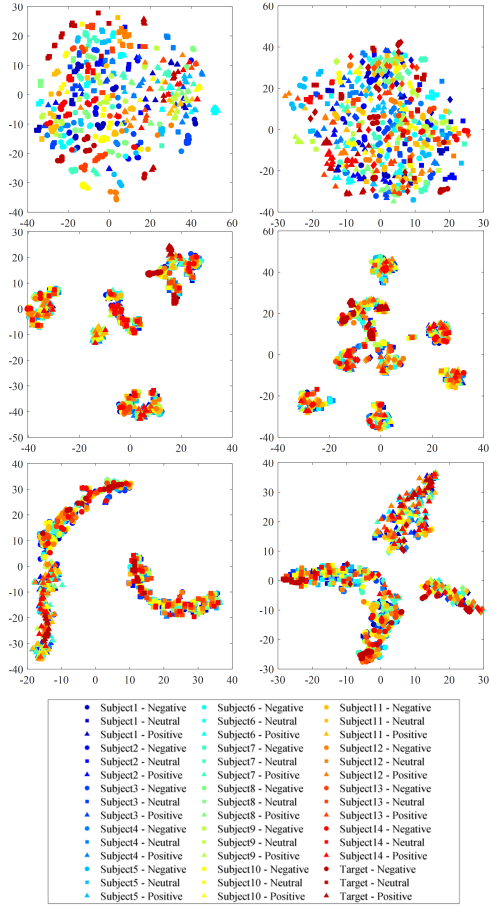

**Figure 6.** The t-SNE visualization of the SEED and SEED-IV datasets at different stages.

/ SS-weighted sources ranked by the SS module. For each S, models are trained with identical settings across seeds, and we report mean±std and the paired improvement Δ Acc. As shown in Table 6, consistent with the previous subsection's observation that SEED exhibits smaller cross-subject differences than SEED-IV, Top-S / SS-weighted yields positive ΔAcc on both datasets but with larger gains on SEED-IV. This indicates that SSAS benefits more when distribution shift is greater or data quality is more heterogeneous, while the gains are correspondingly smaller when inter-subject differences are mild.

### 4.4. Ablation study

We removed individual modules and evaluated the performance of SSAS on the SEED and SEED-IV datasets, as shown in Table. 7. It is worth noting that we selected session 1 of the SEED dataset and session 2 of the SEED-IV dataset for ablation experiments because these two sessions performed the best. The results of the SSAS method after removing various modules on session 1 of the SEED dataset are shown in Fig. 8.

The MDC loss balances the adversarial relationship between domain separability and class separability, preventing premature blurring of domain distinctions. Removing MDC

**Table 6**
Effect of the number of source domains. Random/Equal-w means randomly selecting the source domain (with equal weights), while Top-S/Weighted means selecting the source domain with the highest ranking after SS. Best results **bolded**.

**(a) SEED**

| Setting | Accuracy (%) | | Δ Acc |
| --- | --- | --- | --- |
| | Random / Equal-w | Top-S / SS-weighted | |
| S = 2 | 73.82 ± 12.25 | **74.10 ± 6.90** | +0.28 |
| S = 5 | 83.10 ± 11.80 | **84.25 ± 7.13** | +1.15 |
| S = 10 | 86.49 ± 7.19 | **88.95 ± 7.85** | +2.46 |
| S = 15 | 87.36 ± 6.95 | **91.97 ± 7.82** | +4.61 |

**(b) SEED-IV**

| Setting | Accuracy (%) | | Δ Acc |
| --- | --- | --- | --- |
| | Random / Equal-w | Top-S / SS-weighted | |
| S = 2 | 49.15 ± 13.25 | **55.66 ± 10.90** | +6.51 |
| S = 5 | 60.40 ± 13.97 | **66.92 ± 13.10** | +6.52 |
| S = 10 | 69.85 ± 9.13 | **75.63 ± 11.82** | +5.78 |
| S = 15 | 70.97 ± 10.95 | **77.99 ± 10.56** | +7.02 |

**Table 7**
Ablation Experiment Results.

| Dataset | Method | Accuracy | Change |
| --- | --- | --- | --- |
| SEED | Ours full | 91.97% | |
| | w/o MDC loss | 78.61% | −13.36% |
| | w/o MMD loss | 76.83% | −15.14% |
| | w/o Adversarial | 86.63% | −5.34% |
| | w/o SS | 87.36% | −4.61% |
| | w/o Noise | 90.41% | −1.56% |
| SEED-IV | Ours full | 77.99% | |
| | w/o MDC loss | 74.02% | −3.97% |
| | w/o MMD loss | 66.78% | −11.21% |
| | w/o Adversarial | 72.97% | −5.02% |
| | w/o SS | 70.97% | −7.02% |
| | w/o Noise | 76.04% | −1.95% |

increased the Supremum of $\min_{S \in \mathbb{O}} d_{\mathcal{H} \Delta \mathcal{H}}(S, T)$ (proposition 1) and led to a performance drop of 13.36% on SEED and 3.97% on SEED-IV, highlighting its role in enhancing generalization. Removing the MMD loss significantly weakened generalization, with SEED and SEED-IV accuracies dropping by 15.14% and 11.21%, respectively. This confirms that reducing inter-subject distributional differences is crucial for cross-subject emotion recognition. "w/o adversarial" led to a 5.34% and 5.02% accuracy drop on SEED and SEED-IV, respectively, demonstrating their effectiveness in mitigating inter-subject differences. "w/o Noise" led to a 1.56% and 1.95% accuracy drop on SEED and SEED-IV, respectively, demonstrating that Gaussian noise can effectively improve the model's generalization ability. Finally, removing SS resulted in a 4.61% and 7.02% accuracy drop on SEED and SEED-IV. The SS method contributed more to SEED-IV, likely due to its ability to filter out source domains with poor transferability in complex scenarios.

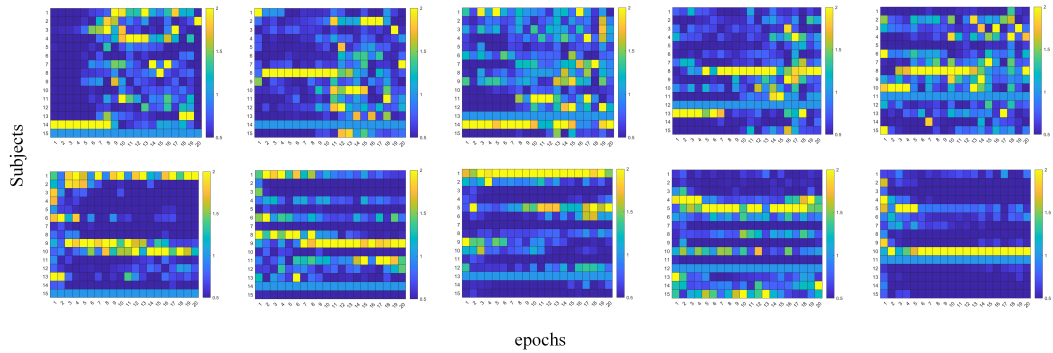

**Figure 7.** The impact of epochs on the weight matrix for both the SEED and SEED-IV datasets. The first row displays the source domain weight variation diagram for the top 5 subjects of the SEED dataset after 20 epochs, while the second row depicts the same for the top 5 subjects of the SEED-IV dataset.

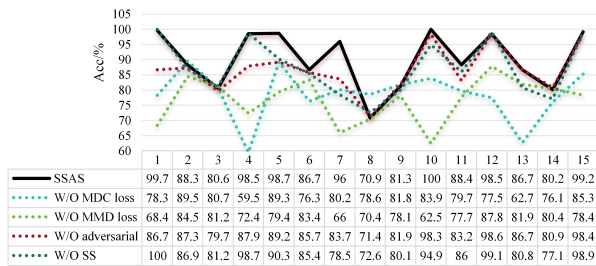

**Figure 8.** The ablation experiment results of the SSAS on session 1 of the SEED dataset

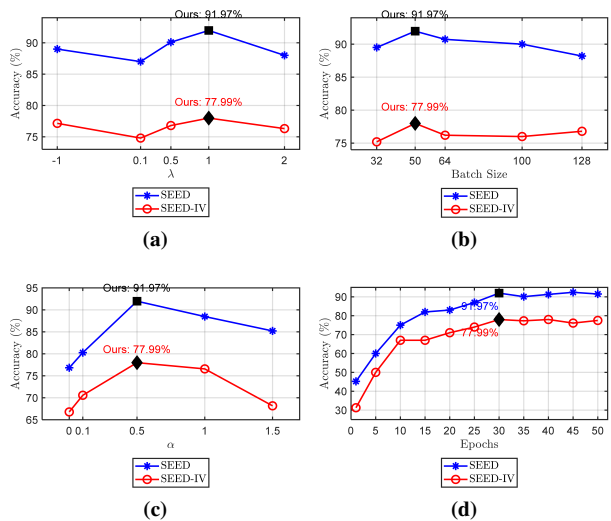

**Figure 9.** Hyperparameter sensitivity analysis on the SEED and SEED-IV datasets.

### 4.5. Hyperparameter sensitivity analysis

A total of 3 hyperparameters are employed in SSAS, which include the batch size, the weight $\alpha$ in the SSAS

loss function, and $\lambda$ in the GRL. The ranges of these hyperparameters are as follows: $\lambda \in -1, 0.1, 0.5, 1, 2$, batch size $\in 32, 50, 64, 100, 128$, and $\alpha \in 0, 0.1, 0.5, 1, 1.5$. The hyperparameters sensitivity analysis results are shown in Fig. 9. Additionally, we analyzed the relationship between accuracy and the training epochs, with the selected SSAS hyperparameters highlighted in black. The experimental results demonstrate that SSAS achieves stable performance in a wide range, verifying its robustness.

### 4.6. Model complexity analysis

The SSAS does not directly perform task learning, instead, it first conducts source domain selection through the SS module before entering the main training phase. To assess the computational efficiency of the SSAS, we analyze the model complexity of both the SS and AS modules. As shown in Table 8, detailed information regarding the training time, number of parameters, and model size for each module is presented. The network architectures used in the SS and AS modules are similar. However, the AS modules introduces an additional domain discriminator to prevent the model from losing domain discriminability during early training. Consequently, the AS module exhibits slightly higher parameter counts and computational complexity than the SS modules.

## 5. Conclusion

We presented SSAS, a domain adaptation framework that couples a learnable source-selection stage with adversarial adaptation. The core idea is a reverse-simulation mechanism that uses domain labels to quantify source transferability and select source domains most relevant to the target domain. SSAS then performs dual adaptation—adversarial learning to reduce inter-domain separability and MMD minimization for cross-subject alignment—while a $\mathcal{L}_{mdc}$ constrains the discrepancy between discriminators across the two stages, preventing premature domain confusion. Theoretical analysis tightens the error bound, and extensive experiments demonstrate consistent gains, especially when low-quality sources are present.

**Table 8**
Ablation experiment of network architecture.

| Network | Complexity | SEED | | | | SEED-IV | | | |
|---------|-----------|------|--------|-------|------|------|--------|-------|------|
| | | Time | Params | FLOPs | Size | Time | Params | FLOPs | Size |
| SS | $\mathcal{O}(d^2 + h(d + o) + 2o)$ | 167.64s | 934.46K | 46.75M | 3.56MB | 79.97s | 240.06K | 12.03M | 0.92MB |
| AS | $\mathcal{O}(d^2 + h(d + o) + 3o)$ | 456.17s | 936.77K | 46.79M | 3.57MB | 190.52s | 241.38K | 12.07M | 0.93MB |

**Strengths vs. related methods.** Unlike SOTA adversarial or pseudo-labeling methods (e.g., DANN-MAT, SH-MDA, DANN-RPLI), SSAS estimates source transferability prior to adaptation, improving robustness to negative transfer and stabilizing gains under high source heterogeneity.

**Limitations.** When domain discrepancies are small and sources are uniformly strong, the trainable selection stage adds computation with limited marginal benefit. Future work will explore lighter selection or adaptive gating to retain SSAS's robustness while reducing overhead in near-homogeneous settings.

## Acknowledgments

This work was supported in part by the Ministry of Higher Education Malaysia under the Fundamental Research Grant Scheme (FRGS) [Grant No. FRGS-EC/1/2024/ICT02/UNIMAP/02/8].

**Table 9**
Additional Notations for Proofs

| symbol | definition |
|---|---|
| $S$ | Source domain |
| $T$ | Target domain |
| $\mathcal{X}$ | Instance set or input space mapping |
| $f$ | Ground truth function |
| $h$ | Prediction function |
| $\epsilon_s(h)$ | Source error of $h$ |
| $\epsilon_T(h)$ | Target error of $h$ |
| $I(h)$ | Indicator function |
| $\lambda_e, \lambda_1', \lambda_2'$ | Minimum joint error rate of $\mathcal{H}$ |

**Basic knowledge**

To provide a comprehensive demonstration of the proof process, Table 9 presents some notations used in this paper.

There are two assumptions on source distribution $\mathbb{O}$: the ground truth function $f$ and hypotheses $h$. The average difference between $h$ and $f$ is defined as:

$$\epsilon_s(h, f) = \mathbf{E}_{x \sim \mathbb{O}}[|h(x) - f(x)|] \quad (15)$$

Given two probability distributions, $S$ and $T$, on the input space $\mathcal{X}$, where $\mathcal{H}$ represents the hypothesis space on $\mathcal{X}$, and $I(h)$ denote the indicator function (i.e., $x \in I(h) \Leftrightarrow h(x) = 1$). Then, the H divergence between $S$ and $T$ is:

$$d_{\mathcal{H}}(S, T) = 2 \sup_{h \in \mathcal{H}} |\Pr_S[I(h)] - \Pr_T[I(h)]| \quad (16)$$

$I(h)$ can be understood as the subset of the input space classified as 1 by $h$, i.e., $I(h) = \{x \mid h(x) = 1\}$. Therefore, $d_{\mathcal{H}}$ is the difference in probabilities of $I(h)$ over distributions $S$ and $T$, where it's noted that sup over all $h \in \mathcal{H}$, meaning selecting the hypothesis $h$ that maximizes this probability difference.

For a hypothesis space $\mathcal{H}$, its corresponding $\mathcal{H}\Delta\mathcal{H}$ space is:

$$g \in \mathcal{H}\Delta\mathcal{H} \Leftrightarrow g(x) = h(x) \oplus h'(x) \text{ for some } h, h' \in \mathcal{H} \quad (17)$$

For a set of prediction functions $\mathcal{H}$, for any $h \in \mathcal{H}$, the following inequality holds with at least a probability of $1 - \delta$:

$$\epsilon_T(h) \leq \epsilon_S(h) + \frac{1}{2} d_{\mathcal{H}\Delta\mathcal{H}}(S, T) + \lambda_e \quad (18)$$

$d_{\mathcal{H}\Delta\mathcal{H}}(S, T)$ represents the distribution discrepancy between the source domain and the target domain, which is also associated with $\mathcal{H}$. $\lambda_e$ represents the minimum joint error rate of $\mathcal{H}$ on these different domains, which actually implies the relationship between the distributions of these different domains, while also being related to $\mathcal{H}$.

**Proof of proposition 1.**

According to Eq. 11, we have

$$\epsilon_T(h) \leq \epsilon_S(h) + \frac{1}{2} d_{\mathcal{H}\Delta\mathcal{H}}(S, T) + \lambda_1', \forall h \in \mathcal{H}, \forall S \in \mathbb{O}. \quad (19)$$

This further yields the Supremum of $\epsilon_S(h)$, we have

$$\epsilon_S(h) \leq \epsilon_{\sum_i^L \varphi_i S_i}(h) + \frac{1}{2} d_{\mathcal{H}\Delta\mathcal{H}}\left(\sum_i^L \varphi_i S_i, S\right) + \lambda_2', \forall h \in \mathcal{H}. \quad (20)$$

Since $\epsilon_{\sum_i^L \varphi_i S_i}(h) = \sum_i^L \varphi_i \epsilon_{S_i}(h)$
and $d_{\mathcal{H}\Delta\mathcal{H}}\left(\sum_i^L \varphi_i S_i, S\right) \leq \max_{j,k} d_{\mathcal{H}\Delta\mathcal{H}}(S_j, S_k)$
we have:

$$\epsilon_T(h) \leq \sum_i^L \varphi_i \epsilon_{S_i}(h) + \frac{1}{2} d_{\mathcal{H}\Delta\mathcal{H}}(S, T) + \frac{1}{2} d_{\mathcal{H}\Delta\mathcal{H}}\left(\sum_i^L \varphi_i S_i, S\right) + \lambda', \forall h \in \mathcal{H}, \forall S \in \mathbb{O} \quad (21)$$

$$\epsilon_T(h) \leq \sum_i^L \varphi_i \epsilon_{S_i}(h) + \frac{1}{2} d_{\mathcal{H}\Delta\mathcal{H}}(S, T) + \frac{1}{2} \max_{j,k} d_{\mathcal{H}\Delta\mathcal{H}}(S_j, S_k) + \lambda', \forall h \in \mathcal{H}, \forall S \in \mathbb{O} \quad (22)$$

where $\lambda' = \lambda_1' + \lambda_2'$. Eq. 22 for all $S \in \mathbb{O}$ holds. Therefore, we complete the proof of bound relating the multi-source and target error.

Why is it that $d_{\mathcal{H}\Delta\mathcal{H}}\left(\sum_i^L \varphi_i S_i, S\right) \leq \max_{j,k} d_{\mathcal{H}\Delta\mathcal{H}}(S_j, S_k)$ holds? Meaning indicated: The divergence of the differences between the overall multi-source domain and the single target domain is constrained by the pair $S_j$ and $S_k$ in space $\mathcal{H}\Delta\mathcal{H}$. $\max_{j,k} d_{\mathcal{H}\Delta\mathcal{H}}(S_j, S_k)$ represents the pair of hypotheses that maximizes divergence, providing an upper bound for the overall domain.)

$\sum_i^L \varphi_i \epsilon_{S_i}(h)$ is the weighted average from multi-source domains. The distance from this weighted average distribution to any single distribution will not exceed the maximum distance between the source distributions. In more detail, $d_{\mathcal{H}\Delta\mathcal{H}}(S_j, S_k)$ represents the maximum divergence value among all pairs of hypotheses within the hypothesis space $\mathcal{H}\Delta\mathcal{H}$, while $d_{\mathcal{H}\Delta\mathcal{H}}\left(\sum_i^L \varphi_i S_i, S\right)$ denotes the maximum divergence value among a finite number of pairs of hypotheses within the hypothesis space $\mathcal{H}\Delta\mathcal{H}$.

Thus, $d_{\mathcal{H}\Delta\mathcal{H}}\left(\sum_i^L \varphi_i S_i, S\right) \leq \max_{j,k} d_{\mathcal{H}\Delta\mathcal{H}}(S_j, S_k)$ holds.